\documentclass{llncs}

\usepackage{microtype}
\usepackage{amsmath,amssymb}
\usepackage{xspace}
\usepackage{booktabs}
\usepackage{graphicx}
\usepackage{multirow}
\usepackage[svgnames,dvipsnames]{xcolor}
\usepackage{subcaption}      
\usepackage{siunitx}
\usepackage{array}
\usepackage{tabularx}
\newcommand{\fscore}[1][1]{$F_{#1}$\xspace}

\begin{document}

\title{Boosting Transformers for Job Expression Extraction and Classification \\ in a Low-Resource Setting}
\titlerunning{Boosting Transformers in a Low-Resource Setting}

\author{Lukas Lange\inst{1,2,3}
	\and Heike Adel\inst{1}
	\and Jannik Str\"otgen\inst{1}
}

\institute{	    
	Bosch Center for Artificial Intelligence \\Robert-Bosch-Campus 1, 71272 Renningen, Germany \\
	\email{\{Lukas.Lange,Heike.Adel,Jannik.Stroetgen\}@de.bosch.com} \\
	\and Spoken Language Systems (LSV),
	\and Saarbr\"ucken Graduate School of Computer Science \\
	Saarland Informatics Campus, Saarland University, Saarbr\"ucken, Germany
}
\maketitle
{\let\thefootnote\relax\footnotetext{\textit{IberLEF 2021, September 2021, Málaga, Spain.}\\Copyright \textcopyright\ 2021 for this paper by its authors. Use permitted under Creative Commons License Attribution 4.0 International (CC BY 4.0).}}
\pagenumbering{gobble}
\pagestyle{empty}

\begin{abstract}
In this paper, we explore possible improvements of transformer models in a low-resource setting. In particular, we present our approaches to tackle the first two of three subtasks of the MEDDOPROF competition, i.e., the extraction and classification of job expressions in Spanish clinical texts.
As neither language nor domain experts, we experiment with the multilingual XLM-R transformer model and tackle these low-resource information extraction tasks as sequence-labeling problems. 
We explore domain- and language-adaptive pretraining, transfer learning and strategic datasplits to boost the transformer model. Our results show strong improvements using these methods by up to 5.3 \fscore points compared to a fine-tuned XLM-R model. Our best models achieve 83.2 and 79.3 \fscore for the first two tasks, respectively. 
\end{abstract}

\begin{keywords}
  Named Entity Recognition
   \and Neural Sequence Tagging 
   \and Domain- and Language-adapted Language Models 
   \and Strategic Datasplits 
\end{keywords}

\section{Introduction}

Information extraction in non-standard domains is a challenging problem due to the large number of complex terms and unusual document structures \cite{friedrich-etal-2020-sofc}. Despite this, pretrained transformer models demonstrated robustness across languages and domains. However,  these models still show their best performance when applied to targets similar to their pretraining corpora which can limit their applicability in many situations \cite{gururangan-etal-2020-dont}.  
One example for this is the Spanish clinical domain, where both, language and domain, can be considered a non-standard setting in the English-centric NLP community \cite{marimon2019automatic}. 

In this paper, we explore possible enhancements of transformer models to overcome this domain and language gap in the context of the MEDDOPROF shared task~\cite{meddoprof-overview}. 
In particular, we participate in the first two tasks of MEDDOPROF \cite{meddoprof-overview}, a challenge concerned with the extraction, classification and normalization of job-related expressions in Spanish clinical texts.  The first task \textit{NER} requires the extraction of three different kinds of occupation and the second task \textit{CLASS} demands to classify each of the previously extracted occupations into four classes reflecting the holder of that job.

We approach this challenge as \textsc{\textbf{N}either \textbf{L}anguage \textbf{N}or \textbf{D}omain \textbf{E}xperts} (NLNDE)
and model them as sequence labeling tasks.
Our solution for these tasks is a neural sequence tagger based on multilingual transformer models. 
In particular, we experiment with continuing the masked language modeling pretraining of the multilingual XLM-R model~\cite{conneau-etal-2020-unsupervised} on Spanish texts, transferring trained models between the two tasks \cite{lange2021share} and using strategic datasplits \cite{wecker-etal-2020-clusterdatasplit}.  

Our results highlight the importance of domain- and language-adapted transformer models, as well as the advantages of combining several models trained on challenging datasplits with ensembling techniques.
Using these methods, our best models achieve \fscore-scores of 83.2 and 79.3 for the two tasks and outperform a fine-tuned XLM-R model by 4.2 and 5.3 \fscore points, respectively.

\section{Related Work}
The MEDODPROF challenge follows a series of shared tasks on Spanish clinical information extraction, including the MEDDOCAN shared task on medical document anonymization \cite{marimon2019automatic} and the PharmaCoNER shared task on concept extraction \cite{gonzalez-agirre-etal-2019-pharmaconer,lange-etal-2019-nlnde-pharmaconer}. 
Main findings of all of these challenges were that transformer models become more commonly used \cite{gonzalez-agirre-etal-2019-pharmaconer,marimon2019automatic} as they begun to dominate the field of information extraction due to their general applicability across languages and domains.
For an overview of recent approaches to low-resource NLP, we refer the refer to \cite{hedderich-etal-2021-survey}.

As the inclusion of domain knowledge via domain-specific embeddings in these special settings is often beneficial \cite{friedrich-etal-2020-sofc,lange-etal-2019-nlnde-meddocan}, we explore domain- and language-adaptive pretraining of transformer models in this paper.
Several recent works have shown that this kind of adaptation boosts performance for downstream tasks in non-standard domains by, e.g.,
pretraining with masked language modeling (MLM) objectives on documents from the target domain \cite{beltagy-etal-2019-scibert,gururangan-etal-2020-dont}. 

In addition, we analyze the effects of model transfer between the first tasks of the challenge, as model transfer between related tasks in similar domains can result in significant performance gains \cite{lange2021share}. 

Further, there is a line of work now questioning traditional train-dev splits \cite{gorman-bedrick-2019-need} as well as random splits \cite{sogaard-etal-2021-need}. More challenging datasplits can be created by clustering the documents based on their similarity, where each split encodes unique information to a certain degree
\cite{wecker-etal-2020-clusterdatasplit}. We use this method to train ensembles of models on these splits in a cross-validation format, such that each model has observed slightly different training instances.

\begin{figure}
  \centering
  \includegraphics[width=\linewidth]{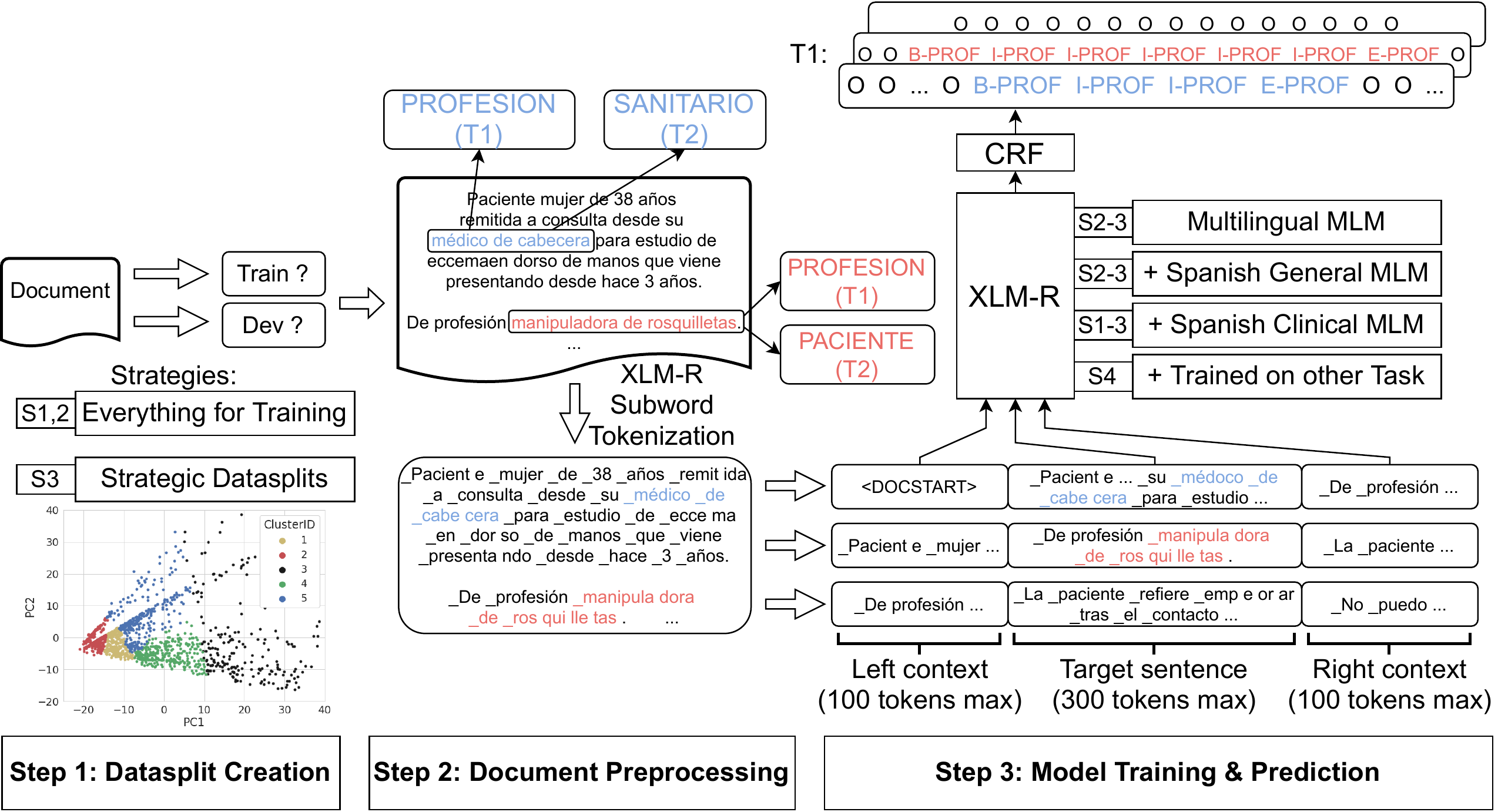}
  \caption{Overview of the NLNDE system architecture. We mark the system variants S1-S4 referring to our different submissions.    
   T1 and T2 refer to task 1 and task 2 of the shared task, respectively.
}
  \label{fig:system}
\end{figure}

\section{Approach}
This section provides an overview of the different methods we used for the two tasks. The complete overview of our system is given in Figure 1. 

\newcommand{\subA}{\textit{S1}\xspace}
\newcommand{\subB}{\textit{S2}\xspace}
\newcommand{\subC}{\textit{S3}\xspace}
\newcommand{\subD}{\textit{S4}\xspace}
\newcommand{\subE}{\textit{S5}\xspace}

\subsection{Document Preprocessing}\label{sec:prep}
Tokenization can be challenging in non-standard domains, including the clinical domain \cite{lange2020cantemist}. We thus use the XLM-R subword tokenizer and perform sequence labeling on the subtoken level with spacy for sentence segmentation. Initial experiments showed possible improvements of up to 2 \fscore points compared to NER on token level.

\subsection{Domain- and Language-specific Masked Language Modeling}
We use XLM-R \cite{conneau-etal-2020-unsupervised} as the main component of our models. XLM-R is a pretrained multilingual transformer model for 100 languages, including Spanish. It shows superior performance in different tasks across languages, and can even outperform monolingual models in certain settings. 
It was pretrained on a large-scale corpus, and Spanish documents made up only 2\% of this data, as provided in Table 1. 

Thus, we explore further pretraining of this model and tune it towards Spanish documents by pretraining either on (1) a medium-size Spanish corpus with general domain documents \cite{jose_canete_2019_3247731} or (2) a smaller Spanish clinical corpus consisting of the MeSpEN resources \cite{villegas2018mespen} and publicly available Scielo articles. 
Note that the clinical corpus was not part of the general-domain corpus.

We use masked language modeling for pretraining and trained for three epochs over the corpora, which roughly corresponds to 30k steps for the smaller clinical corpus and 685k steps for the general-domain corpus using a batch-size of 4.
Finally, we have three XLM-R variants that we compare in this paper:

\begin{enumerate}
    \item Standard \textbf{XLM-R} pretrained on 100 languages by \cite{conneau-etal-2020-unsupervised}. 
    \item \textbf{Spanish XLM-R} based on standard XLM-R with further pretraining using Spanish documents from the general domain.
    \item Spanish \textbf{Clinical XLM-R} based on standard XLM-R with further pretraining using Spanish documents from the clinical domain.
\end{enumerate}

\begin{table}[t]
    \begin{minipage}{.5\linewidth}
        \caption{Sizes of Pretraining Corpora.}
        \centering
        \begin{tabular}{ll} \toprule
        Corpus & Size \\ \midrule
        Original XLM-R corpus   & 2.4 TB \\
        -- Spanish Subcorpus    & 53.3 GB \\ \midrule
        Spanish Corpus          & 17.3 GB \\
        Spanish Clinical Corpus & 789.2 MB \\ \bottomrule
        \end{tabular}
        \label{tab:sizes}
    \end{minipage}
    \begin{minipage}{.5\linewidth}
        \caption{Hyperparameters.}
        \centering
        \begin{tabular}{ll} \toprule
        \multicolumn{2}{c}{\textit{XLM-R}} \\
        Embedding size & 1024      \\
        Max. sentence length & 300 \\
        Context to left/right & 100    \\ \midrule
        \multicolumn{2}{c}{\textit{Optimizer}} \\
        Batch size            & 16 \\
        Learning rate         & $2e-5$ \\
        $\beta_1$, $\beta_2$  & 0.9, 0.999 \\
        Weight decay          & 0.0 \\\midrule
        \multicolumn{2}{c}{\textit{Training}} \\
        Epochs   & 20 \\ 
        Early stopping & training loss \\
        & \textit{or} on dev. set \\
        \bottomrule
        \end{tabular}
        \label{tab:hyper}
    \end{minipage}
\end{table}

\subsection{Sequence Tagger}
For the sequence tagger, we use one of the XLM-R models, either the standard XLM-R or one of our adapted models, and apply a CRF layer on top \cite{lafferty-etal-2001-crf}.
We add this CRF layer to address the problem of longer multi-word annotations, as job descriptions often span several tokens or, in our case, subtokens (as explained in Section 3.1). In addition, a CRF prevents inconsistencies in the labels. 

We split all sentences to a maximum length of 300 subtokens and add the context of up to 100 subtokens to the left/right to get cross-sentence information. 
The labels are in BIOSE encoding, which is an extended BIO encoding with additional labels for the last token of an annotation (E-) and single-token annotations (S-)

Our model architecture is basically the same across all runs. We only exchange the transformer model. The models are trained using an AdamW optimizer for a maximum of 20 epochs. Our hyperparameters are given in Table 2.

\subsection{Strategic Datasplits}
We test two options to train the sequence taggers: 

(1) Using all of the available training data and stop training according to the training loss. This method  provides the model with the most input instances. However, the stopping criterion is not as meaningful as using the task's metric on a held-out validation set. 

(2) Thus, as our second method, we split the data into train and validation sets. Then, we train the model using only the train-fraction of all the data and use the held-out validation data to determine the best model, which is then used to annotate the test data. 
As an alternative to random splits, we follow \cite{wecker-etal-2020-clusterdatasplit} and create strategic datasplits by clustering the documents according to their similarity. This creates more challenging splits, as more distant documents are left out for validation. 

For this, each document is represented as the average vector of the XLM-R embeddings for each token. This document representation is reduced to five dimensions using PCA. Finally, the documents are clustered into five equally-sized splits using $k$-Means clustering. 
We train five models for each task and embedding with each having a different validation split. Our splits are visualized in Figure 2. We see that clusters 1, 2 and 4 are densely populated with highly similar documents, while clusters 3 and 5 contain more distinct documents.

To better understand the strategic datasplits, we analyzed whether the different medical topics included in the corpus correlate with the splits. However, we found that the strategic clusters incorporate more diverse information than just topic similarity as there is no substantial overlap between topics and datasplits.

\begin{figure}
    \centering
    \includegraphics[width=0.6\linewidth]{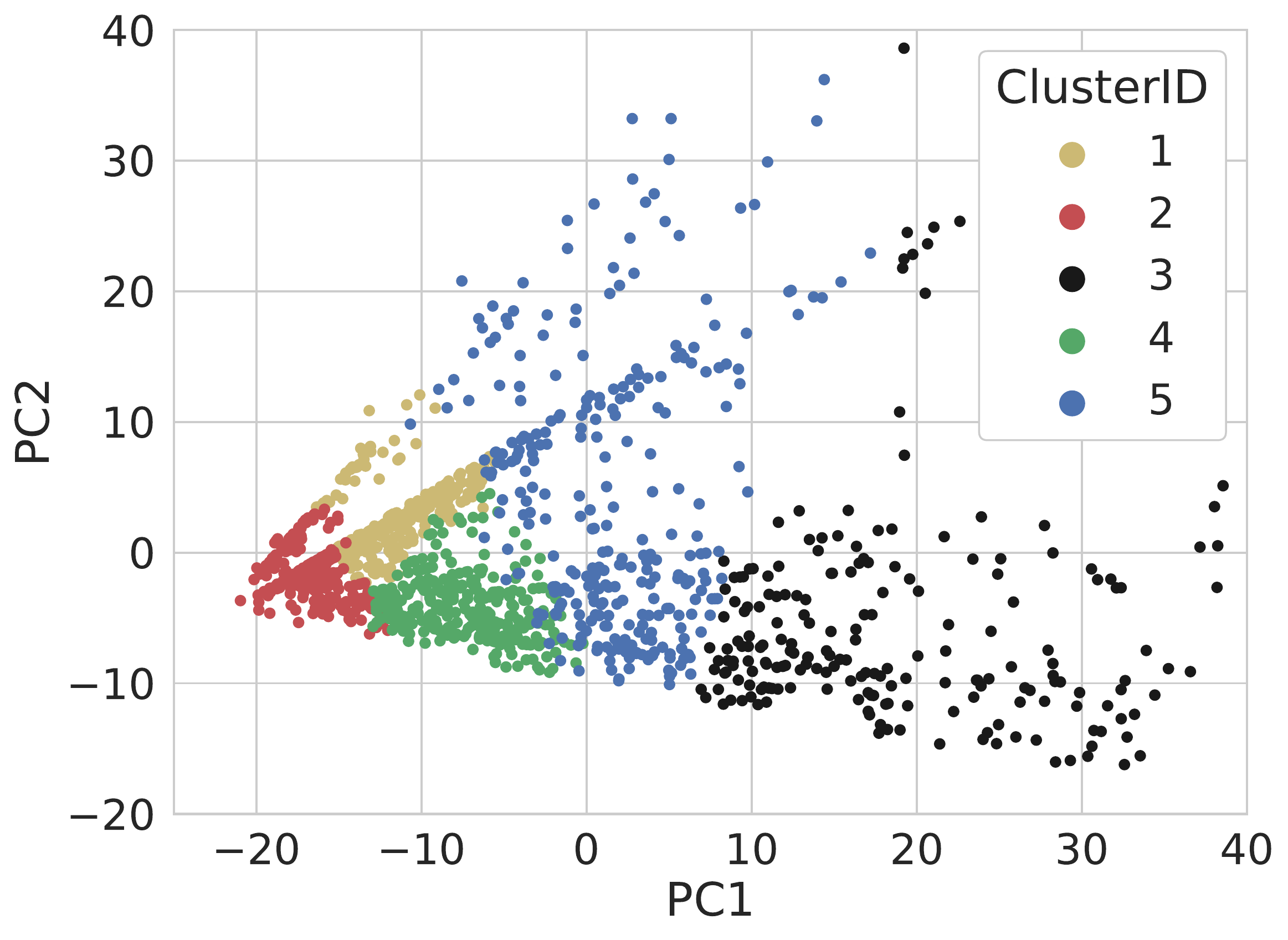}
    \caption{Our strategic datasplits in the two-dimensional space with PCA. }
    \label{fig:splits}
\end{figure}

\subsection{Ensembling of Model Predictions}
In order to capture the different advantages of multiple models, we combine them using ensembling. This is particularly helpful when models carry different types of information. For example, the models trained using our strategic datasplits all have seen a slightly different training set, and, thus, combining them using ensembling should further improve performance. 
We apply ensembling by majority voting. For this, we use hard voting that counts the labels by each model and does not consider the CRF probabilities. We convert the BIOSE labels to BIO labels for the ensembling process as the simpler BIO encoding leads to fewer conflicts. 
Further, we apply postprocessing of the label sequence to correct inconsistencies in the resulting label sequence and enforce for 0.23\% of predictions for the test set that the first token of each annotations begins with \textit{B-}.

\subsection{Transfer Learning}
As the first two tasks of the challenge are related and can possibly benefit from each other, we explore the potential of model transfer between them. For example, having basic knowledge of what and which occupations (task 1) are mentioned in a text can be useful to determine whether occupations are related to the patient or to someone else (task 2).
For this, we first train models on the auxiliary task and then transfer the resulting model to the targeted main task. In our case, the auxiliary task is either task 1 or 2 and the other task is the main task.

\subsection{Submissions}
The following five runs are the NLNDE submissions to the MEDDOPROF shared task. We use the same model architectures for both tasks. 
Note that all submissions, except for \subA are ensembles of multiple models based on the three different embeddings. 
In Section~\ref{sec:results}, we compare these submissions with further model variations.

\begin{itemize}
    \item [\subA]: The Spanish clinical XLM-R model trained on the complete training data. 
    \item [\subB]: All three XLM-R language models combined in one ensemble (3 models).
    \item [\subC]: Ensemble of models trained using strategic datasplits (15 models). 
    \item [\subD]: Ensemble of models based on transfer learning from the auxiliary task to the main task (3 models).
    \item [\subE]: The combination of all above models into one ensemble (21 models).
\end{itemize}

\section{Results}\label{sec:results}
Our official results for the first two tasks of the MEDDOPROF shared task are given in Table 3.
In addition, we include several other models to analyze the performance of each embedding, because most of our submissions are ensembles combining all three XLM-R embeddings. 
The official evaluation metric is the \fscore-score and our best models are highlighted.

We find that the Spanish XLM-R trained on general-domain Spanish data is often the best transformer compared to the standard XLM-R and the clinical one, probably because it was trained on the largest amount of data. In addition, the extraction of occupations is not unique to the clinical domain and general-domain Spanish knowledge seems to be beneficial for this as well. 

We find that model transfer (\subD) is only useful when transferring models from task 1 (the detection of occupations) to task 2 (the classification of occupations). Reusing models that already learned the detection of profession expressions as an auxiliary task improves the main task, e.g., by up to 2.9 \fscore points for XLM-R. 

Training models on strategic datasplits (\subC) provides the best results overall, and is even better than the ensemble of all models (\subE). The strategic datasplits improved the ensemble model \subB by 0.4 and 2.3 \fscore points for task 1 and 2. Note that this submission unintentionally contained the Spanish general-domain XLM-R models with transfer learning in the ensemble. These were trained without our strategic datasplits. The corrected results are marked with "*".

The overall best model for task 1 is the ensemble of general-domain XLM-R trained using the strategic datasplits with an \fscore-score of 83.2. This model was not submitted as a run to the shared task, but shows the importance of the language-adaptive pretraining and the usefulness of strategic datasplits.

\definecolor{Gray}{gray}{0.85}
\newcolumntype{a}{>{\columncolor{Gray}}c}
\newcommand{\bestS}[1]{\underline{#1}}
\newcommand{\best}[1]{\textbf{#1}}
\newcommand{\bestB}[1]{\underline{\textbf{#1}}}

\begin{table}
\caption{Results on the official test set. Our best submission is highlighted with underlines and the overall best model in bold. }
\centering
\begin{tabular}{l|ccc|ccc} \toprule
 & \multicolumn{3}{c}{Task 1} & \multicolumn{3}{|c}{Task 2} \\
 & Precision & Recall & F1 & Precision & Recall & F1 \\ \midrule
\multicolumn{7}{c}{\textit{trained on all training instances}} \\
XLM-R                         & 83.9 & 75.7 & 79.6 & 76.9 & 71.3 & 74.0 \\
\textbf{clinical XLM-R (S1)}  & 82.5 & 76.2 & 79.2 & 81.6 & 74.0 & 77.6 \\
Spanish XLM-R                 & 84.5 & 76.9 & 80.5 & 80.2 & 74.6 & 77.3 \\
\textbf{ensemble of all (S2)} & 85.1 & 77.7 & 81.2 & 80.6 & 73.7 & 77.0 \\ \midrule
\multicolumn{7}{c}{\textit{transfer training}} \\
XLM-R                         & 82.5 & 75.2 & 78.7 & 80.8 & 73.4 & 76.9 \\
clinical XLM-R                & 81.7 & 75.2 & 78.3 & 81.0 & 72.6 & 76.6 \\
Spanish XLM-R (TSX)           & 81.6 & 75.0 & 78.2 & 81.5 & 74.9 & 78.1 \\
\textbf{ensemble of all (S4)} & 83.8 & 76.6 & 80.0 & 81.9 & 74.3 & 77.9 \\ \midrule 
\multicolumn{7}{c}{\textit{trained on strategic datasplits}} \\
XLM-R                         & 83.9 & 75.0 & 79.2 & 81.2 & 74.3 & 77.6 \\
clinical XLM-R                & 83.8 & 76.6 & 80.0 & 80.7 & 75.4 & 77.9  \\
Spanish XLM-R                 & \best{86.3} & \best{80.4} & \best{83.2} & 82.5 & \best{75.9} & 79.1  \\
ensemble of all * & \best{86.3} & 78.7 & 82.3 & 82.1 & 75.5 & 78.7 \\
\textbf{ensemble of all} + TSX \textbf{(S3)} & \bestS{85.5} & \bestS{78.3} & \bestS{81.8} & \bestB{83.0} & \bestB{75.9} & \bestB{79.3} \\ \midrule \midrule
\textbf{ensemble of all models (S5)} & 84.7 & \bestS{78.3} & 81.4 & \bestB{83.0} & 75.7 & 79.2 \\ \bottomrule
\end{tabular}
\label{tab:res-ner}
\end{table}

\section{Conclusion}
In this paper, we described our submissions for the first two tasks of the MEDDOPROF competition. By utilizing domain- and language-adaptive pretraining, strategic datasplits and ensembling methods, we were able to improve already high-performing transformer-based models by up to 5.3 \fscore points and achieved competitive results in the competition as neither language nor domain experts. Future work will include the exploration of different clinical corpora with our newly trained Spanish XLM-R models. 

\section*{Acknowledgments}
The authors would like to thank the anonymous reviewer for the helpful comments and the text mining group at the barcelona supercomputing center for the smooth organization of MEDDOPROF.

\bibliographystyle{splncs04}
\bibliography{references}

\end{document}